\documentclass[runningheads]{llncs}
\usepackage{graphicx}

\usepackage{tikz}
\usepackage{comment}
\usepackage{amsmath,amssymb} 
\usepackage{color}

\usepackage{epsfig}
\usepackage{graphicx}
\usepackage{algorithm}
\usepackage{algpseudocode}
\renewcommand{\Comment}[2][.5\linewidth]{%
  \leavevmode\hfill\makebox[#1][l]{//~#2}}

\usepackage{algorithmicx} 
\usepackage{hyperref}
\usepackage{adjustbox}
\usepackage{subfig}
\usepackage{booktabs}

\newcommand{\beq}{\begin{equation}}
\newcommand{\eeq}{\end{equation}}
\newcommand{\etal}{{et al}.\@ }

\begin{document}
\pagestyle{headings}
\mainmatter
\def\ECCVSubNumber{2841}  

\title{Key Frame Proposal Network for Efficient Pose Estimation in Videos} 

\titlerunning{Key Frame Proposal Network for Efficient Pose Estimation in Videos}
%

\author{Yuexi Zhang\inst{1} \orcidID{0000-0001-5012-5459}\and
Yin Wang\inst{2} \orcidID{0000-0001-6810-0962}\and \\
Octavia Camps\inst{1} \orcidID{0000-0003-1945-9172} \and
Mario Sznaier \inst{1} \orcidID{0000-0003-4439-3988}}
\authorrunning{Y.Zhang, Y.Wang, O.Camps and M.Sznaier}
%
\institute{Electrical and Computer Engineering, Northeastern University, Boston, MA 02115  \\
\email{zhang.yuex@northeastern.edu, camps,msznaier@coe.neu.edu}\\
\url{http://robustsystems.coe.neu.edu/} \and
Motorola Solutions, Inc., Somerville, MA 02145\\
\email{yin.wang@motorolasolutions.com}}

\maketitle
\begin{abstract}
Human pose estimation in video  relies on local information by either estimating each frame independently or tracking  poses across frames. In this paper, we propose a novel method combining local approaches with global context. We introduce a light weighted, unsupervised, key frame proposal network (K-FPN) to select informative frames and a learned  dictionary to recover the entire pose sequence from these frames. The K-FPN speeds up the pose estimation  and provides robustness to  bad frames with occlusion, motion blur, and illumination changes, while the learned dictionary provides global dynamic context.  Experiments on Penn Action and sub-JHMDB datasets show that the proposed method achieves state-of-the-art  accuracy, with substantial speed-up.

\end{abstract}

\begin{figure*}[h]
\centering
\includegraphics[width=\linewidth]{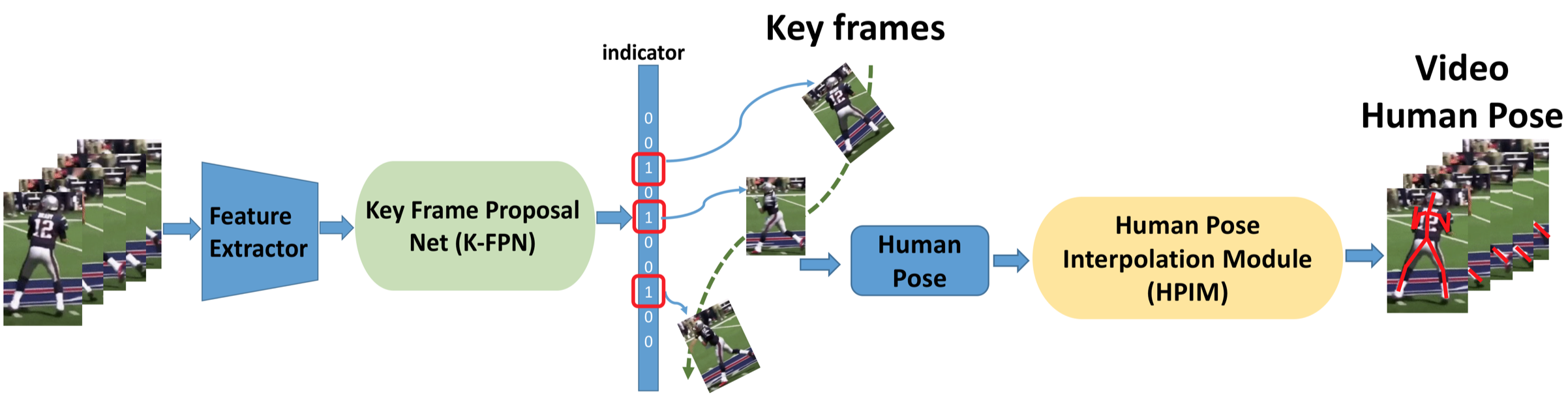}
  \caption{\textbf{Proposed pipeline for video human pose detection.} The K-FPN net, which is trained  unsupervised,   selects a set of key frames. The Human Pose Interpolation Module (HPIM), trained to learn  human pose dynamics,   generates  human poses for the entire input sequence from the poses in the key frames.}
\label{fig:visual}
\end{figure*}

\section{Introduction}

Human pose estimation \cite{belagiannis2017recurrent,Hourglass,pishchulin2013strong,deeppose,wei2016convolutional}, which seeks to estimate the locations of  human body joints, has many practical applications such as smart video surveillance \cite{cristani2013human,park2008understanding}, human computer interaction \cite{shotton2011real}, and VR/AR\cite{lin2010augmented}.

The most general pose estimation pipeline extracts features from the input, and then uses a classification/regression model to predict the location of the joints. 
Recently, \cite{sparsely-labeled} introduced a Pose Warper capable of  using a few manually annotated frames to propagate pose information across  the complete video. 
However,  it relies on annotations of every $k^{th}$ frame and thus it fails to fully exploit  the dynamic correlation between them.

\begin{figure}[h!]
\centering
\includegraphics[width=0.9\linewidth]{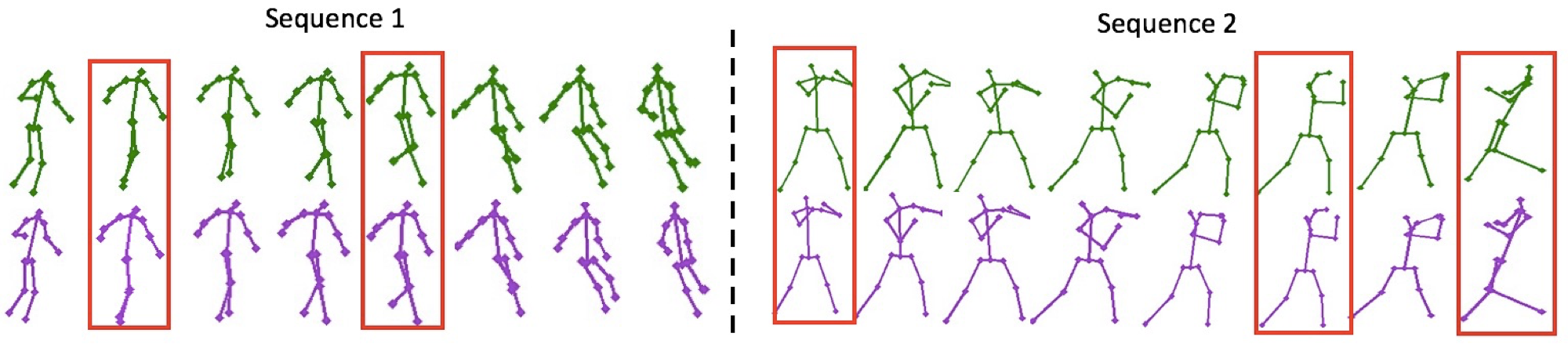}
  \caption{Two examples of the output of our pipeline.  Top: ground truth. Bottom: poses recovered from the automatically selected key frames (red boxes).}
\label{fig:caption}
\end{figure}

Here, we propose an alternative pose estimation pipeline based on  two observations: All frames are not equally informative; and the dynamics of the body joints  can be modeled using simple dynamics.   
The new pipeline uses a light weighted {\em key frame proposal} network (K-FPN), shown in Fig.~\ref{fig:visual}, to  select  a small number of frames to apply a pose estimation model. One of the main contributions of this paper is  a new  loss function based on the {\em recovery error in the latent feature space} for {\em  unsupervised training} of this network. The second module of the pipeline is an efficient Human Pose Interpolation Module (HPIM), which uses a dynamics-based dictionary to  obtain the  pose in the remaining frames. 
Fig.~\ref{fig:caption} shows two sample outputs of our pipeline, where the poses shown in purple were interpolated from the automatically selected red key frames.
The advantages of the proposed approach are:
\begin{itemize}
    \item It uses a very light, unsupervised, model to  select ``important" frames.
    \item It is highly efficient, since  pose is estimated only  at key frames.
    \item It is robust to challenging conditions present in the non-key frames, such as occlusion, poor lighting conditions, motion blur, etc.
    \item It can be used to reduce annotation efforts for supervised approaches by  selecting which frames should be manually annotated.
\end{itemize}   

\section{Related Work}
\noindent {\bf Image Based Pose Estimation.}
Classical approaches use the  structure and  inter-connectivity among the body parts and rely on hand-crafted features. Currently, deep networks are used instead of hand-crafted features.  \cite{chen2014articulated} used  Deep Convolutional Neural Networks (DCNNs) to learn the conditional probabilities for the presence of parts and their spatial relationships.  \cite{yang2016end} combined in an end-to-end framework the DCNN with the expressive mixture of parts model.    
\cite{chu2016structured} learned the correlations among body joints using an ImageNet pre-trained VGG-16 base model. \cite{wei2016convolutional}  implicitly modeled long-range dependencies  for  articulated pose estimation.   \cite{Hourglass} 
proposed a  ``hourglass" architecture to handle  large pixel displacements, opening a pathway to incorporate  different scaled features stacked together.   \cite{flownet,featurepriamid,PapandreouZKTTB17,Tang_dlcm,yang2017pyramid} made several improvements on multi-scaled feature pyramids for estimating human pose.  
\textcolor{black}{However, capturing sufficient scaled features is computationally expensive.  \cite{fastpose} proposed a teacher-student architecture to reduce network complexity and  computational time. Finally, \cite{openpose,PiPaf,nie2018mula} refined the location of keypoints by exploiting the human body structure.}

\noindent{\bf Video Based Pose Estimation.}  Human pose estimation can be improved by capturing temporal and appearance information across frames.  
\cite{simonyan2014two,song2017thin} use deep Convolutional Networks (ConvNet) with optical flow as its input motion features.  \cite{pfister2015flowing} shows that an additional convolutional layer is able to learn a simpler model of the spatial human layout.  \cite{charles2016personalizing} improves this work to demonstrate that the joint estimations can be propagated from poses on the first few frames by integrating optical flows. Furthermore, tracking on poses is another popular methodology such as \cite{poseTrack,simple-baseline} which can jointly refine estimations. Others adopt Recurrent Neural Networks(RNN) \cite{luo2018lstm,gkioxari2016chained,3Dlstm}. \cite{gkioxari2016chained} shows that a sequence-to-sequence model can work for structured output prediction. A similar work  \cite{luo2018lstm} imposes sequential geometric consistency to handle image quality degradation. Despite of notable accuracy, RNN-based methods suffer from the expensive computations required.  \cite{nie2019dynamic} proposed to address this issue by using a light-weighted distillator to online distill pose kernels by leveraging the temporal information among frames.

\section{Proposed Approach}
Fig.\ref{fig:visual} shows the proposed architecture. Given $T$ consecutive frames, we aim to select a small number of  frames, which can capture the global context and provide enough information to interpolate the poses in the entire video. This is challenging  since  annotations for this task are usually unavailable.  Next, we  formulate this problem  as the minimization of a  loss function, which allows us to provide a set of optimal proposals deterministically and without supervision.

The main intuition behind the proposed architecture is that there is a high degree of spatial and temporal correlation in the data, which can be captured by  a simple dynamics-based model. Then, key frames should be selected such that they are enough (but no more than strictly needed) to learn the dynamic model and recover the non-selected frames.

\subsection{Atomic Dynamics-based Representation of Temporal Data}

We will represent the dynamics  of the input data by using the  dynamics-based atomic (DYAN) autoencoder introduced in \cite{DYAN}, where the atoms are the   
 impulse response  $
y(k) =c p^{k-1}
$ of  linear time invariant  (LTI) systems  with a  pole\footnote{Poles are in general complex numbers.  Systems with real outputs  with a non real pole $p$ must also have its conjugate pole $p^*$: $y(k) = c p^{k-1} + c^*.{p^*}^{(k-1)}$.} $p$,
 $c$ is a constant and  $k$ indicates time. The model uses $N \gg T$  atoms, collected as columns of a dictionary matrix
$\mathbf{D} \in \mathbb{R}^{T\times N}$:
\beq
\mathbf{D} = \left [ \begin{array}{cccc}
1 & 1 & \dots & 1 \\
p_1 & p_2 & \dots & p_N \\
\vdots & \vdots & \vdots & \vdots \\
p_1^{T-1} & p_2^{T-1} & \dots & p_N^{T-1} \end{array} \right ]
\label{eq:D}
\eeq

Let $\mathbf{Y} \in  \mathbb{R}^{T\times M}$ be the input data matrix, where  each column  has the temporal evolution of a datapoint (i.e.  one coordinate of a human joint or  the value of a feature, from time 1 to time $T$). 
Then, we represent $\mathbf{Y}$  by a  matrix $\mathbf{C} \in \mathbb{R}^{N\times M}$ such that
$
{\mathbf{Y}} = \mathbf{D}\mathbf{C}
$,
where  the element $\mathbf{C}(i,j)$ indicates how much of the output of the $i^{th}$ atom  is used to recover the $j^{th}$ input data  in $\mathbf{Y}$:
\[
\mathbf{Y}(k,j)  = \sum_{i=1}^N  \mathbf{C}(i,j)p_i^{k-1}
\]
In \cite{DYAN}, the dictionary $\mathbf{D}$ was learned  
from training data to predict future frames by minimizing a loss function that penalized the reconstruction error of the input and the $\ell_1$ norm of $\mathbf{C}$ to  promote the sparsity of $\mathbf{C}$ (i.e. using as few atoms/pixel as possible):
\beq
\mathcal{L}_{dyn} = \|\mathbf{Y} - \mathbf{DC}\|^2_2 + \alpha \sum_{i,j}|\mathbf{C}(i,j)| \label{eq:Ldyn}
\eeq
In this paper, we propose a different loss function to learn $\mathbf{D}$, which is better suited to the task of  key frame selection. Furthermore, the learning procedure in \cite{DYAN} requires solving a Lasso  optimization problem  for each input before it can evaluate the loss (\ref{eq:Ldyn}).  In contrast, the loss function we derive in section \ref{sec:loss} is computationally  very efficient, since it does not require such optimization step.

\subsection{Key frame Selection Unsupervised Loss}\label{sec:loss}
\label{subsect:loss}

Given an input video $\mathcal{V}$ with $T$ frames, consider a tensor of its deep features $\mathcal{Y} \in \mathbb{R}^{T\times c \times w \times h}$ with $c$ channels of width $w$ and height $h$, reshaped into a matrix $\mathbf{Y} \in \mathbb{R}^{T\times M}$. That is, the element $\mathbf{Y}(k,j)$ has the value of the feature $j$, $j=1,\dots,M = cwh$, at time $k$. 
Then, our goal is to select a subset of key frames, as small as possible, that captures the content of all the frames.  Thus, we propose to cast this problem as  finding a {\em minimal subset of rows} of $\mathbf{Y}$ (the \emph{key frames}), such that it would be possible to recover the left out frames (the other rows of $\mathbf{Y}$) by using these few frames and their atomic dynamics-based representation.

\begin{problem}\label{p:KF}
{Given a matrix of features $\mathbf{Y} \in \mathbb{R}^{T\times M}$,  an overcomplete dictionary $\mathbf{D} \in \mathbb{R}^{T \times N}$, $N \gg T$,  for which there exist an  atomic dynamics-based representation $\mathbf{C} \in \mathbb{R}^{N \times M}$ such that $\mathbf{Y} = \mathbf{D} \mathbf{C}$, find a binary selection matrix $\mathbf{P}_r \in \mathbb{R}^{r \times T}$  with the least number of rows $r$, such that 
$\mathbf{Y} \approx \mathbf{D} \mathbf{C}_r$, where $\mathbf{C}_r \in \mathbb{R}^{N\times M}$ is the atomic dynamics-based representation of  the selected key frames $\mathbf{Y}_r = \mathbf{P}_r\mathbf{Y}$.}
\end{problem}

Problem~\ref{p:KF} can be written as the following optimization problem: 
\begin{eqnarray}
&& \min_{r,\mathbf{P}_r \in \mathbb{R}^{r \times T}}  \|\mathbf{Y} - \mathbf{D}\mathbf{C}_r\|_F^2 + \lambda r, \label{eq:obj}\\
&& \text{subject to:} \nonumber\\
&& \mathbf{P}_r \mathbf{Y}= \mathbf{P}_r \mathbf{D}\mathbf{C}_r  \label{eq:KFC1} \\ 
&&\mathbf{P}_r(i,j) \in \{0,1\} \ \ 
\sum_j \mathbf{P}_r(i,j) = 1\ \ 
\sum_i \mathbf{P}_r(i,j) \le 1 \label{eq:KFC2}
\end{eqnarray}
The first term in the objective (\ref{eq:obj}) minimizes the recovery error while the second term penalizes the number of frames selected. The  constraint (\ref{eq:KFC1}) establishes that $\mathbf{C}_r$ should be the atomic dynamics-based representation of the  key frames and the constraints (\ref{eq:KFC2}) force the binary selection matrix $\mathbf{}P_r$  to  select $r$ distinct frames. However, this problem is hard to solve since the optimization variables are  integer ($r$) or  binary  (elements of $\mathbf{P}_r$).

Next, we show how we can obtain a relaxation of this problem, which is  differentiable  and suitable as a unsupervised loss function to train our key frame proposal network. {\color{black} The derivation has three main steps. First, we use the constraint (\ref{eq:KFC1}) to replace $\mathbf{C}_r$ with an expression that depends on $\mathbf{P}_r$, $\mathbf{D}$ and $\mathbf{Y}$. Next, we  make a change of variables so we do not have to minimize with respect to a matrix of unknown dimensions.  Finally, in the last step we relax the constraint on the binary variables  to be real between 0 and 1.}

\noindent {\bf Eliminating $\mathbf{C}_r$:} Consider the atomic dynamics-based representation of $\mathbf{Y}$:
\begin{equation}
\mathbf{Y} = \mathbf{D}\mathbf{C} \label{eq:DC}
\end{equation}
Multiplying both sides by $\mathbf{P}_r$, defining $\mathbf{D}_r = \mathbf{P}_r\mathbf{D}$, and using (\ref{eq:KFC1}), we have:
\beq
\mathbf{P}_r\mathbf{Y} = \mathbf{D}_r\mathbf{C} = \mathbf{D}_r\mathbf{C}_r \label{eq:Cr0}
\eeq
 Noting that $\mathbf{D}_r$ is an overcomplete dictionary, we select the solution for  $\mathbf{C}_r$ from (\ref{eq:Cr0}) with  minimum Frobenious norm, which can be found by solving:
 \beq
 \min_{\mathbf{C}_r} \|\mathbf{C}_r\|^2_F  \ \ \text{subject to: } \mathbf{P}_r \mathbf{Y}= \mathbf{D}_r\mathbf{C}_r
 \eeq
 The solution of this problem is:
 \beq
 \mathbf{C}_r = \mathbf{D}_r^T(\mathbf{D}_r\mathbf{D}_r^T)^{-1}\mathbf{P}_r\mathbf{Y}
 \label{eq:Cr} 
 \eeq
 since the rows of $\mathbf{D}$ (see (\ref{eq:D})) are linearly independent and hence the inverse $(\mathbf{D}_r\mathbf{D}_r^T)^{-1}$ exists. 
{\color{black} Substituting \eqref{eq:Cr} in the first term in  \eqref{eq:obj} we have: 
\beq
\|\mathbf{Y} - \mathbf{D}\mathbf{C}_r\|_F^2 = \left\|[ \mathbf{I}  - \mathbf{DD}_r^T( \mathbf{D}_r\mathbf{D}_r^T)^{-1}\mathbf{P}_r] \mathbf{Y}\right\|_F^2 
\eeq
Using the fact that $\mathbf{D}_r=\mathbf{P}_r\mathbf{D}$ yields  the following equivalent to Problem 1:
\beq
\min_{r,\mathbf{P}_r \in \mathbb{R}^{r \times T}} 
 \left\|[ \mathbf{I}  - \mathbf{DD}^T\mathbf{P}_r^T( \mathbf{P}_r\mathbf{D} \mathbf{D}^T\mathbf{P}_r^T)^{-1}\mathbf{P}_r] \mathbf{Y}\right\|_F^2 
 + \lambda r,\  \ \text{subject to }  (\ref{eq:KFC2}) \label{eq:opt1}
\eeq
{\bf Minimizing with respect to a fixed size matrix:} Minimizing with respect to $\mathbf{P}_r$ is difficult because one of its dimensions is $r$, which is  a variable that we also want to minimize. To avoid this issue, we introduce an approximation trick, where we add a small perturbation $\rho > 0$ to the diagonal of  $\mathbf{P}_r\mathbf{D} \mathbf{D}^T\mathbf{P}_r^T$:
\beq
\min_{r,\mathbf{P}_r \in \mathbb{R}^{r \times T}}  \left\|[\mathbf{I}  - \mathbf{DD}^T\mathbf{P}_r^T(\rho\mathbf{I}+ \mathbf{P}_r\mathbf{D} \mathbf{D}^T\mathbf{P}_r^T)^{-1}\mathbf{P}_r] \mathbf{Y}\right\|_F^2 + \lambda r,\  \ \text{subject to }  (\ref{eq:KFC2}) \label{eq:opt2}
\eeq
and combine (\ref{eq:opt2}) with the Woodbury matrix identity 
$$\mathbf{A}^{-1}-\mathbf{A}^{-1}\mathbf{U}[\mathbf{B}^{-1}+\mathbf{VA}^{-1}\mathbf{U}]^{-1}\mathbf{VA}^{-1}=
[\mathbf{A}+\mathbf{UBV}]^{-1}$$
by setting $\mathbf{A} =  \mathbf{I}$, $\mathbf{U} =\mathbf{DD}^T \mathbf{P}_r^T$, $\mathbf{B}^{-1} =\rho\mathbf{I}$, and $\mathbf{V}=\mathbf{P}_r$, to get:
\beq
\min_{r,\mathbf{P}_r \in \mathbb{R}^{r \times T}} 
\left \|[\mathbf{I} + \rho^{-1} \mathbf{DD}^T\mathbf{P}_r^T\mathbf{P}_r]^{-1}\mathbf{Y}\right\|_F^2 + \lambda r,\  \ \text{subject to }  (\ref{eq:KFC2}) \label{eq:opt3}
\eeq
Now, define  $\mathbf{S}= \mathbf{P}_r^T\mathbf{P}_r$, which is a matrix of fixed size $T \times T$.  Furthermore using the constraints (\ref{eq:KFC2}), it is easy to show that $\mathbf{S}$ is diagonal and that its diagonal elements $\mathbf{s}_{i}$ are 1 if $\mathbf{P}_r$ selects frame $i$ and 0 otherwise. Thus,  the vector $\mathbf{s} =  \text{diagonal}(\mathbf{S})$ is  an indicator vector for the sought  key frames and the number of key frames is given by  $r = \sum_i \mathbf{s}_i$. Therefore, the objective becomes:
\beq
\min_{\mathbf{s} \in \mathbb{R}^{T \times 1}, \mathbf{s}_i \in \{0,1\}} 
\left \|[\mathbf{I} + \rho^{-1} \mathbf{D}\mathbf{D}^T{\mathbf{S}}]^{-1}\mathbf{Y}\right\|_F^2 + \lambda \sum_i \mathbf{s}_i  \label{eq:opt4}
\eeq
Note that the fact that the inverse $(\mathbf{I} + \rho^{-1} \mathbf{D}\mathbf{D}^T\text{diagonal}(\mathbf{s})]^{-1}$ is well defined follows from Woodbury's identity and the fact that $(\rho\mathbf{I}+\mathbf{P}_r\mathbf{D} \mathbf{D}^T\mathbf{P}_r^T)^{-1}$ exists since $\rho>0$ and $\mathbf{P}_r\mathbf{D} \mathbf{D}^T\mathbf{P}_r^T$ is positive semi-definite.

\noindent{\bf Relaxing the binary constraints:} Finally, we relax the binary constraints on the elements of the indicator vector $\mathbf{s}$ and let them be real numbers between 0 and 1. We now have the differentiable objective function:
\beq
\min_{\mathbf{s} \in \mathbb{R}^{T \times 1}, 0 \le \mathbf{s}_i \le 1 } 
\left \|[\mathbf{I} + \rho^{-1} \mathbf{D}\mathbf{D}^T{\mathbf{S}}]^{-1}\mathbf{Y}\right \|_F^2 + \lambda \sum_i \mathbf{s}_i  \label{eq:opt5}
\eeq
where the only unknown is  $\mathbf{s}= \text{diagonal}(\mathbf{S})$. Then, we can use the loss function:
\beq
\boxed{\mathcal{L}_{K-FPN} = \left \|[\mathbf{I} + \rho^{-1} \mathbf{D}\mathbf{D}^T{\mathbf{S}}]^{-1}\mathbf{Y}\right \|_F^2 + \lambda \sum_i \mathbf{s}_i } \label{eq:loss}
\eeq
where  the vector $\mathbf{s}$ should be the output of a sigmoid layer in order to push its elements to binary values (See section \ref{sec:implementation} for more details).}

\subsection{Human Pose Interpolation}

Given a video with $T$ frames, let $\mathbf{H}_r \in \mathbb{R}^{r\times 2J}$ be  the 2D coordinates of $J$ human joints for $r$ key frames, $\mathbf{P}_r \in \mathbb{R}^{r \times T}$ be the associated selection matrix,   and $\mathbf{D}^{(h)}$ be a dynamics-based dictionary trained on skeleton sequences using a DYAN  autoencoder \cite{DYAN}.
Then, the Human Pose Interpolation Module (HPIM)  finds
the skeletons  $\mathbf{H} \in \mathbb{R}^{T \times 2J}$ for the entire sequence, which can be efficiently computed. Its expression can be derived as follows. 
First, use the reduced dictionary:
$
\mathbf{D}^{(h)}_r = \mathbf{P}_r\mathbf{D}^{(h)}
$
  and (\ref{eq:Cr}) to compute the minimum Frobenius norm atomic dynamics-based representation for the  key frame skeletons $\mathbf{H}_r$:
$
 \mathbf{C}_r = 
 {\mathbf{D}^{(h)}_r}^T
 (\mathbf{D}^{(h)}_r{\mathbf{D}^{(h)}_r}^T)^{-1}\mathbf{H}_r.
 $
Then, using the complete dictionary $\mathbf{D}^{(h)}$, the entire skeleton sequence  $\mathbf{H} = \mathbf{D}^{(h)}\mathbf{C}_r$ is given by:
\beq
\boxed{\mathbf{H} = (\mathbf{D}^{(h)}{\mathbf{D}^{(h)}}^T) \mathbf{P}_r^T [\mathbf{P}_r(\mathbf{D}^{(h)}{\mathbf{D}^{(h)}}^T)\mathbf{P}_r^T ]^{-1}\mathbf{H}_r}
\label{eq:hpim}
\eeq
where   $\mathbf{D}^{(h)}{\mathbf{D}^{(h)}}^T$  can be computed ahead of time.

\subsection{Architecture, Training, and Inference}\label{sec:implementation}

\begin{figure*}[!th]
\centering
\includegraphics[height=0.3\textwidth,width=0.9\linewidth]{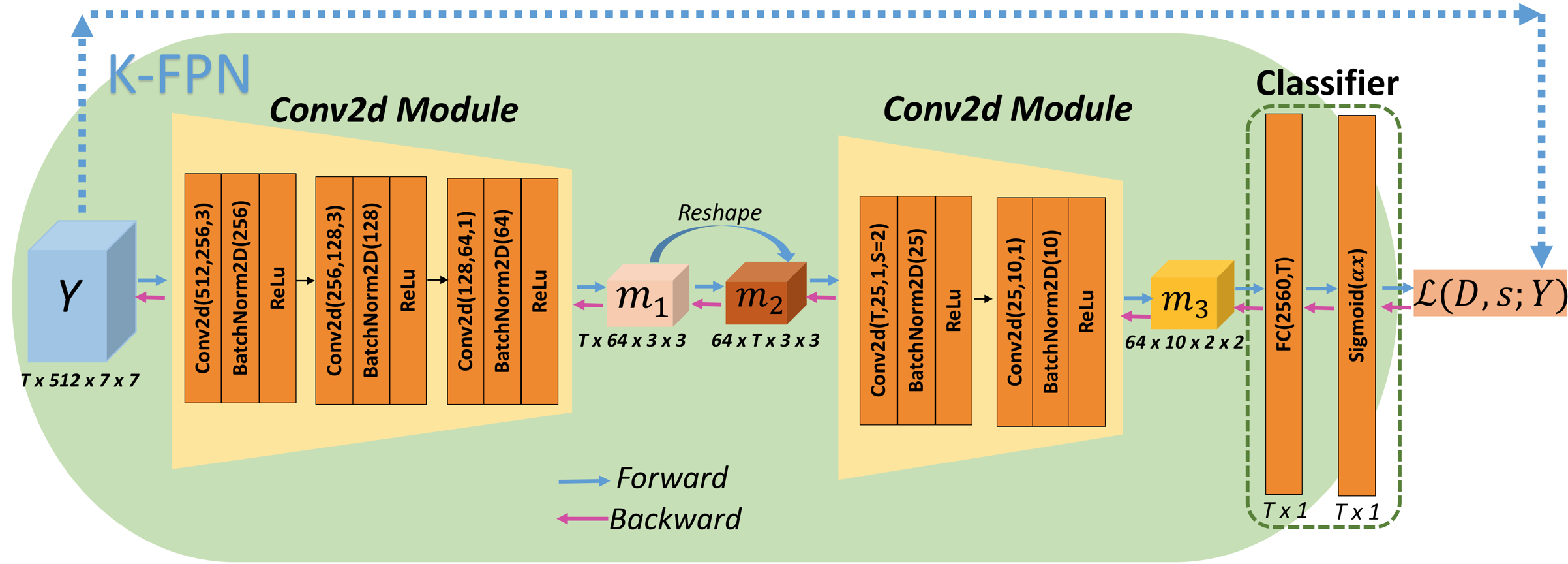}
  \caption{K-FPN Architecture and details of its modules.}
\label{fig:arch}
\end{figure*}
Fig.~\ref{fig:arch} shows  the architecture for the K-FPN, which is trained completely unsupervised, by minimizing the loss (\ref{eq:loss}).   It consists of two Conv2D modules (Conv + BN + Relu) followed by a Fully Connected (FC) and a Sigmoid layers. The first Conv2D   downsizes the input feature tensor from $(T\times512\times7\times7)$ to $(T\times64\times3\times3)$ while the second one uses the temporal dimension as input channels. The $T\times 1$ output of the  FC layer is forced by the Sigmoid layer into logits close to either 0 or 1, where a `1' indicates `key frame' and its index which one.  Inspired by \cite{quantization}, we utilized a control parameter $\alpha$ to form a customized classification layer, represented as $\sigma (\alpha x) = [1 + exp(-\alpha x)]^{-1}$, where 
 $\alpha$ is linearly increased with the training epoch. By controlling  $\alpha$, the output from the K-FPN is nearly a binary indicator such that the sum of its elements is the total number of key frames. 
 The training and inference procedures
 are summarized in 
 Algorithms \ref{algo1} to \ref{algo3}.
 and code is available at \url{https://github.com/Yuexiaoxi10/Key-Frame-Proposal-Network-for-Efficient-Pose-Estimation-in-Videos}.
 
\begin{algorithm}[H]
\caption{Training K-FPN model (Dictionary $\mathbf{D}$)} \label{algo1}
\begin{algorithmic}[1]
    \State \textbf{Input:}Training video sequences $\mathcal{V}$ with up to $T$ frames
    \State \textbf{Output:} key frame indicator $\mathbf{s}$
    \State \textbf{Initialized:}  $\mathbf{D}$ with $N$ poles $p \in \mathbb{C}$ in a ring in  [0.85,1.15] 
    \For{max number of epochs}
        \State $\mathcal{Y} \gets \text{ResNet}(\mathcal{V})$
        \State $m_{1} \gets \text{Conv2D}(\mathcal{Y}) $ \Comment{spatial embedding}
        \State $m_{2} \gets \text{Reshape}(m_{1})$
        \State $m_{3} \gets \text{Conv2D}(m_{2})$ \Comment{temporal embedding}
        \State $\mathbf{F} \gets \text{FC}(m_{3})$ \Comment{mapping to 1D latent space}
        \State $\mathbf{s} \gets \text{Sigmoid}(\mathbf{F})$ \Comment{key frame binary indicator}
        \State Minimize loss $\mathcal{L}_{K-FPN}(\mathbf{D},\mathbf{s};\mathcal{Y})$ \Comment{updating $\mathbf{D}$, $\mathbf{s}$}
    \EndFor
\end{algorithmic}
\end{algorithm}
{ 
\begin{algorithm}[H]
\caption{Training skeleton-based dictionary $\mathbf{D}^{(h)}$ \cite{DYAN} } \label{algo2}
\begin{algorithmic}[1]
    \State \textbf{Input:}Training skeleton sequences $\mathbf{H}$
    \State \textbf{Output:} Atomic Dynamics-based Representation $\mathbf{C}$
    \State \textbf{Initialize:}  $\mathbf{D}^{(h)}$ with poles in a ring $[0.85,1.15] \in \mathbb{C}$
    \For{max number of epochs}
        \State $ \mathbf{C} \gets \text{DYAN}_{\text{encoder}}(\mathbf{H}, \mathbf{D}^{(h)})$
        \State $\hat{\mathbf{H}} \gets \text{DYAN}_\text{decoder}(\mathbf{C}, \mathbf{D}^{(h)})$
        \State Minimize loss $L_{dyn}(\mathbf{H},\hat{\mathbf{H}})$ \Comment{updating $\mathbf{D}^{(h)}$}
    \EndFor
\end{algorithmic}
\end{algorithm}}
\begin{algorithm}[H]
\caption{Inference K-FPN model and Human Pose Interpolation Module} \label{algo3}
\begin{algorithmic}[1]
    \State \textbf{Input:} Testing video sequences $\mathcal{V}$, dictionary $\mathbf{D}^{(h)}$
    \State \textbf{Output:} key frame indicator $\mathbf{s}$, reconstructed human skeletons $\mathbf{H}$ 
    \State $\mathbf{DDT} = \mathbf{D^{(h)}}\mathbf{D^{(h)}}^T$ \Comment{Precompute} 
    \For{all testing sequences}
        \State $\mathbf{s} \gets $ K-FPN($\mathcal{V}$) \Comment{Select Key Frames}
        \State $\mathbf{P}_r \gets \text{SelectionMatrix}(\mathbf{s})$
        \State $\mathbf{H}_r \gets \text{PoseEstimator}(\mathbf{s},\mathcal{V})$ \Comment{key frame skeletons}
        \State $\mathbf{H} = \mathbf{DDT}\cdot \mathbf{P}_r^T \cdot[\mathbf{P}_r \cdot \mathbf{DDT} \cdot \mathbf{P}_r^T ]^{-1}\cdot\mathbf{H}_r$ \Comment{Reconstructed skeletons}
    \EndFor
\end{algorithmic}
 
\end{algorithm}

\begin{figure*}[t!]
\centering
\includegraphics[width=0.9\linewidth]{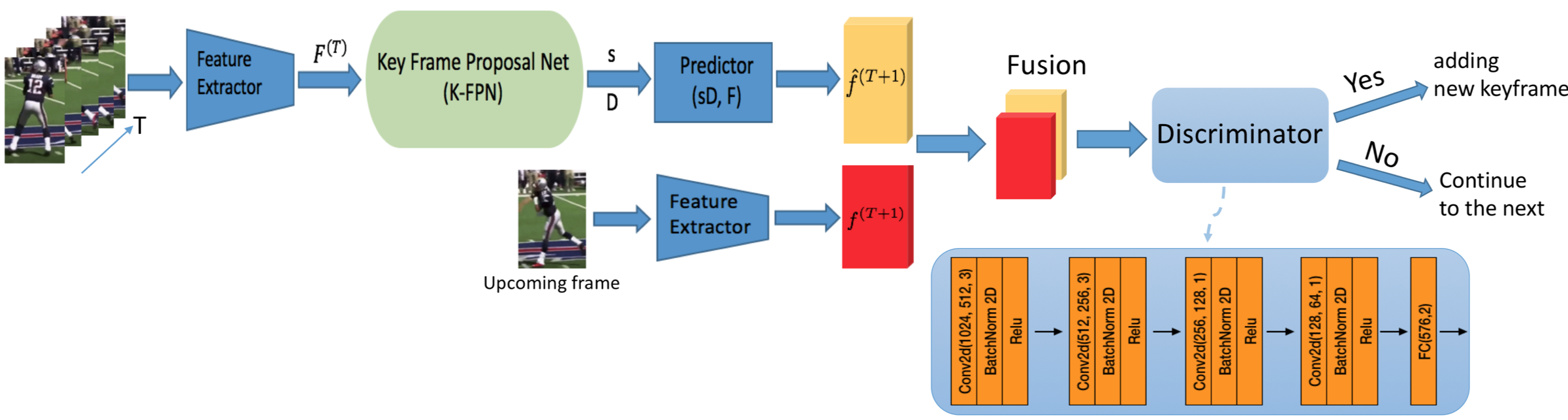}
  \caption{\textbf{Online key frame detection.} The discriminator distinguishes between input features and features predicted from previous key frames to decide if a new frame should be added as a key frame. }
  \label{fig:onlineUpdating}
\end{figure*}

\subsection{Online Key Frame Detection}

The proposed K-FPN can be  modified to process incoming frames, after a minimum set of initial frames has been processed. To do this, we  add a discriminator module as shown in Fig.~\ref{fig:onlineUpdating}, consisting of  four (Conv2D + BN + Relu) blocks, which is used to decide if an incoming frame should be selected as a key frame or not. 
The discriminator is trained to distinguish  between features of the incoming frame  and  features  predicted from the  set of key frames selected so far, which  are easily generated by multiplying the atomic dynamics-based representation of the current key frames with the associated dynamics-based dictionary extended with an additional row (since the number of frames is increased by one)  \cite{DYAN}. The reasoning behind this design is that when the features of the new frame cannot be predicted correctly,  it must be because the frame brings novel information and hence it should be incorporated as a key frame.

\section{Experiments}

Following \cite{luo2018lstm,nie2019dynamic}, we evaluated the K-FPN on two widely-used public datasets: Penn Action \cite{penn_action} and sub-JHMDB \cite{Jhuang:ICCV:2013}. 
Penn Action is a large-scale benchmark, which depicts  human daily activities in unconstrained videos. It has 2326 video clips, with 1258 reserved for training and 1068 for testing with varied frames. It provides 13 annotated joint positions on each frame as well as their visibilities. Following common convention, we only considered the visible joints to evaluate. sub-JHMDB \cite{Jhuang:ICCV:2013} has 319 video clips in three different splits with a training and testing ratio of  roughly 3:1.  It provides 15 annotations  on each human body. However, it only annotates visible joints.  Following  \cite{luo2018lstm,nie2019dynamic,song2017thin}, the evaluation is reported as the average precision over all splits. 

We adopted the ResNet family \cite{Resnet} as our feature encoder and evaluated our method, as the depth was varied from 18 to 101 (see \autoref{subsect:Ablation results}). 
During  training, we froze the ResNet$X$, where $X \in $ [18/34/50/101], and then our K-FPN was trained only on the features output from the encoder. 
Following  \cite{nie2019dynamic}, we  adopted the pre-trained model from \cite{simple-baseline} as our pose estimator. During our experiments, we applied a specific model, which was trained on the MPII\cite{mpii} dataset with ResNet101. However, unlike  previous work \cite{nie2019dynamic}, we did not do any fine-tunning for any of the datasets. To complete the experiments, we split the training set into training and validation parts with a rough ratio of 10:1 and used the validation split to validate our model along with the training process. {The learning rate of K-FPN for both datasets was set as 1e-8 and we used 1e-4 for the online-updating experiment.} The ratio for the two terms in our loss function  (\ref{eq:loss}) is approximately 1:2 for Penn Action and 3:1 for sub-JHMDB.

The K-FPN and HPIM dictionaries were initialized  as  in \cite{DYAN}, with   $T = 40$ rows for both  datasets. Since videos vary in length, we added dummy frames when they had less than 40 frames. For clips  longer than 40 frames, we randomly selected 40 consecutive frames as our input during training  and used an sliding window of size 40 during testing, in order to evaluate the entire input sequence.  

\subsection{Data Preprocessing and Evaluation Metrics}

We followed  conventional data preprocessing strategies. Input images were resized to 3x224x224 and normalized using the parameters provided by \cite{Resnet}. After that, in order to capture a better pose estimation from the pose model, we utilized the person bounding box to crop each image and pad to 384x384 with a varying scaling factor from 0.8 to 1.4. The Penn Action dataset provides such an annotation, while JHMDB does not. Therefore, we generated the person bounding box on each image by using the person mask described in \cite{luo2018lstm}.

Following \cite{nie2019dynamic,luo2018lstm,song2017thin}, we  evaluated our performance using the PCK score  \cite{Yang&Ramanan}: a body joint is considered to be correct only if it falls within a range of ${\beta L}$ pixels, where $L$ is defined by $ L = \max(H,W)$, where $H$ and $W$ denote  the height and width of the person bounding box and ${\beta}$ controls the threshold to justify how precise the estimation is. We follow convention and set ${\beta = 0.2}$.

Our full framework consists of three steps: given an input video of length $T$, K-FPN  first samples $k$ key frames;  then, pose estimation is done on these $k$ frames; and HPIM interpolates these results  for the full sequence. The reported running times are the aggregated time  for these three steps.  
All running times were computed on NVIDIA GTX 1080ti for all methods.

\subsection{Qualitative Examples}

\begin{figure*}[!th]
\centering
\includegraphics[width=\linewidth]{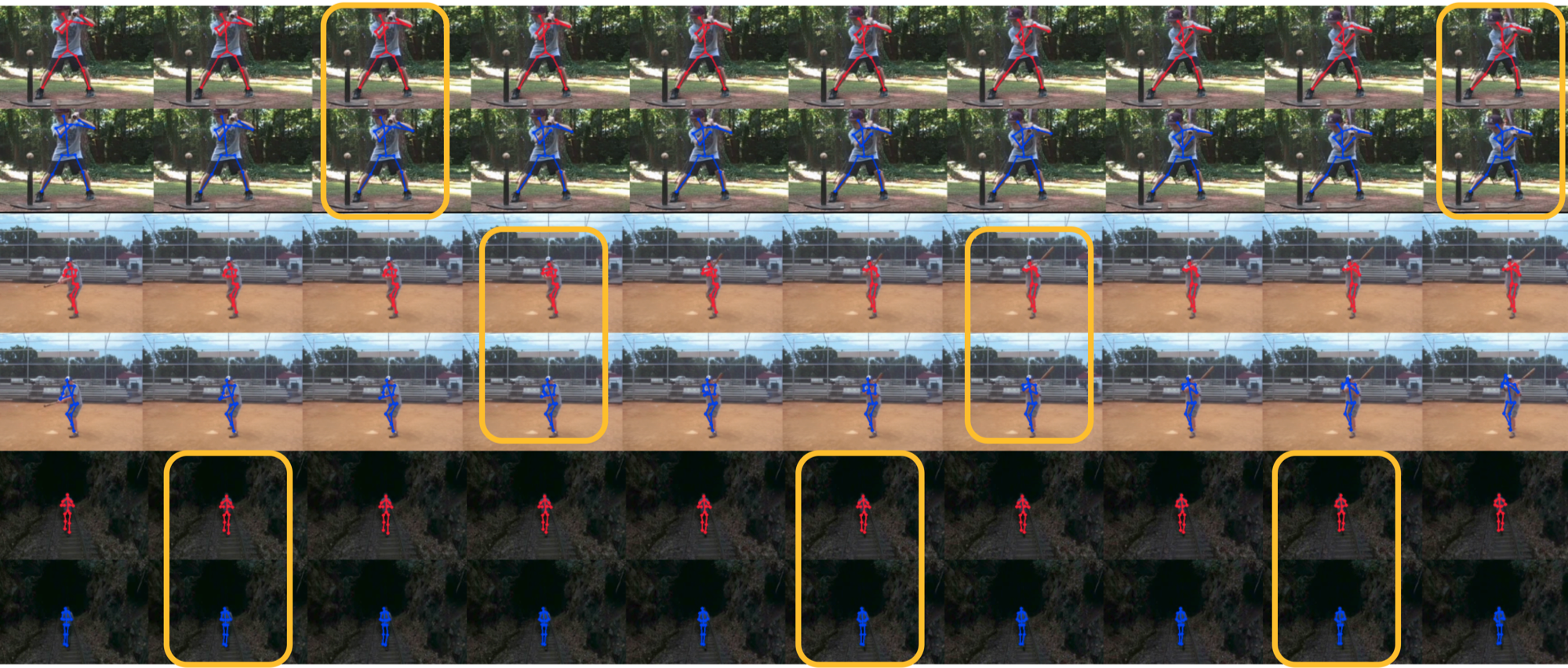}
  \caption{\textbf{Qualitative Examples}. The yellow bounding box indicates key frames chosen by K-FPN. The red skeletons are the ground truth, and blue ones are the ones recovered by the interpolation module HPIM.}
  \label{qualitive}
\end{figure*}

Figs.~\ref{fig:visual} and \ref{qualitive}  show qualitative  examples where it can be seen that the proposed approach can successfully recover the  skeletons  from a few key frames. Please see the supplemental material for more examples and videos.

\subsection{Ablation Studies}
\label{subsect:Ablation results}

In order to evaluate the effectiveness of our approach, we conducted ablation studies on the validation split for each dataset. 
 
 \noindent{\bf Backbone Selection.} We tested K-FPN using different backbones from  the ResNet family. Since sub-JHMDB is not a large dataset, we believe that our K-FPN would be easily overfitted by using deeper feature maps. Thus, we didn't apply ResNet101 on this dataset specifically. 
 Table \ref{ablation study} summarizes the results of this study, where we report running time(ms) and Flops(G) along with PCK scores (higher is better) and average number of selected key frames. These results show that the smaller networks  provide faster speed with minor degradation of the performance. Based on these results, for the remaining experiments we used the best model on the  validation set. Specifically, we used ResNet34 for Penn Action and Resnet18 for sub-JHMDB. 
 
 \begin{table*}[t]
\caption{\textbf{Backbone selection: PCK for sub-JHMDB and Penn Action.}} 
\label{ablation study}
\centering
\begin{adjustbox}{height=0.2\textwidth, width=1.0\linewidth}
\begin{tabular}{|c|c|c|c|c|c|c|c|c|c|c|c|}
\hline
Backbone                                                      & FLOPs(G) & Time(ms) & Head  & Sho.  & Elbow & Wrist & Hip   & Knee  & Ankle & Avg.  & Avg. \#key frames(Std.) \\ \hline
\multicolumn{12}{|c|}{\textbf{Study on sub-JHMDB validation split}}                                                                                                 \\ \hline
\begin{tabular}[c]{@{}c@{}}K-FPN\\ (Resnet50)\end{tabular}   & 5.37     &  6.9    & 98.3 & 98.5 & 97.7 & 95.4 & 98.6 & 98.5 & 98.0 & 97.9  & 17.5(1.5)       \\ \hline
\begin{tabular}[c]{@{}c@{}}K-FPN\\ (Resnet34)\end{tabular}   & 4.68     &  5.7    & 98.0 & 98.3 & 97.3 & 95.4 & 98.2 & 97.8 & 97.2 & 97.5  & 17.1(1.0)      \\ \hline
\begin{tabular}[c]{@{}c@{}}K-FPN\\ (Resnet18)\end{tabular}   & 2.32     &  4.6   & 98.1  & 98.4 & 96.8 & 93.6 & 98.4 & 98.3 & 97.7  & 97.3 & 15.8(1.8)       \\ \hline
\multicolumn{12}{|c|}{\textbf{Study on Penn Action Validation split}}                                                                                                    \\ \hline
\begin{tabular}[c]{@{}c@{}}K-FPN\\ (Resnet 101)\end{tabular} & 10.23    &   9.7   & 99.2 & 98.6 & 97.3  & 95.8  & 98.1  & 97.9 & 97.4  & 97.7 & 17.7(3.1)       \\ \hline
\begin{tabular}[c]{@{}c@{}}K-FPN\\ (Resnet 50)\end{tabular}  & 5.37     &  6.6    & 98.6 & 98.3 & 96.0 & 94.3 & 98.6 & 98.7 & 98.8 & 97.5  & 16.6(4.9)       \\ \hline
\begin{tabular}[c]{@{}c@{}}K-FPN\\ (Resnet 34)\end{tabular}  & 4.68     &   5.5   & 98.2 & 98.1 & 95.1 & 92.9 & 98.5 & 98.7 & 98.6 & 97.1 & 15.0(3.5)      \\ \hline
\end{tabular}
 \end{adjustbox}

\end{table*}

\noindent{\bf Number of Key Frames Selection.} To evaluate the selectivity of the K-FPN, we randomly picked $n=100$ validation instances with $T$ frames,   ran the K-FPN (using Penn action validation set with Resnet34) and recorded the number of key frames selected for each of these instances: $K = [k_{1}, k_{2},...,k_{n}]$.  Given the number of key frames $k_i$, theoretically, one could determine the best selection by evaluating the PCK score for each of the $T \choose k_i$ possibilities.  Since it is infeasible to run that many combinations, we tried two alternatives:  i) selected frames by uniformly sampling the sequence (Uniform Sample), and ii) randomly sampled  100 out of all possible combinations and kept the one with the best PCK score (Best Sample).  Table~\ref{random} compares the average PCK score using the K-FPN against Uniform Sampling and Best Random Sampling.  {\color{black}  From \cite{TempoBai}, it follows that the best PCK score over 100 subsets has a probability $> 95\%$, with $99\%$ confidence, of being the true score over the set of  all possible combinations and hence provides a good estimate of the unknown optimum.  Thus,  our \emph{unsupervised} approach indeed achieves performance very close to the theoretical  optimum.}

\begin{table}[t]
\caption{\textbf{Number of Key frames Evaluation (PCK).} \label{random}
 K-FPN vs best   out of 100 random samples  and uniform sampling on  the Penn Action dataset.  }
\centering
\begin{tabular}{|c|c|c|c|}
\hline
\multicolumn{4}{|c|}{Key frames Selection Method} \\ \hline
        &  K-FPN    &  Best Sample   &  Uniform Sample    \\ \hline
PCK    & 98.0     & 96.4       & 79.3   \\ \hline
\end{tabular}
\end{table}

\begin{table}[]
\caption{\textbf{Online vs Batch Key Frame Selection.} We evaluated the performance on sub-JHMDB using $T = T_b + T_o$ frames.}
\label{table:compareUpdate}

\centering
\begin{adjustbox}{height=0.085\textwidth, width=0.9\linewidth}
\begin{tabular}{|c|c|c|c|c|c|c|c|c|c|}
\hline
\multicolumn{10}{|c|}{\textbf{$\tiny{T_b =30}$, $\tiny{{T_o} = 10}$}}                  \\ \hline
           & Head & Should & Elbow & Wrist & Hip  & Knee & Ankle & Mean & Avg. \#Key frames(Std.) \\ \hline
Online  & 94.8 & 96.3   & 95.2  & 89.6  & 96.7 & 95.2 & 92.3  & 94.4 & 15.2(2.4)     \\ \hline
Batch & 94.7 & 96.3   & 95.2  & 90.2  & 96.4 & 95.5 & 93.2  & 94.5 & 16.3(1.8)     \\ \hline
\end{tabular}
\end{adjustbox}
\label{Online Updating}
\end{table}

\begin{figure*}[!th]
\centering
\includegraphics[height=0.25\textwidth,width=0.8\linewidth]{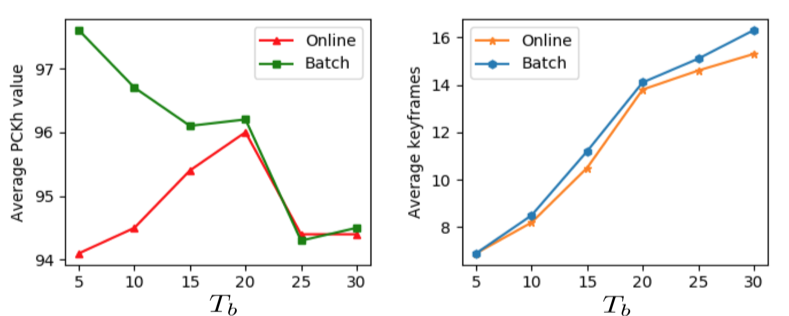}
  \caption{\textbf{Online vs Batch Key Frame Selection.} We evaluated the performance on sub-JHMDB. The entire length of videos to obtain keyframes is $T_b+{T_o}$}
 \label{fig:compareUpdate}
\end{figure*}

\noindent{\bf Online Key Frame Selection.} We compared  the performance between using batch and online updating  key frame selection.  All evaluations were done with the sub-JHMDB dataset. In this experiment, we use a set of $T_b$  frames to select an initial set of key frames (using  ``batch" mode) and process the following $To = 10$ frames using online detection. We compare the achieved PCK score and the number of selected frames against the results obtained using a batch approach on all $T_b + T_o$ frames. The results of this experiment for $T_b=30$ and for $5 \le T_b \le 30$  are shown in Table~\ref{table:compareUpdate} and Fig.~\ref{fig:compareUpdate}, respectively. 
This experiment shows that on one hand, using batch mode, shorter videos ($T_b+T_o$ small) have better PCK score than longer ones. This is  because the beginning of the action is often simple (i.e. there is little motion at the start) and is well represented with very few key frames. On the other hand,  online updating performs as well as batch, as long as the initial set of frames is big enough ($T_b =20$ frames). This can be explained by the fact that if $T_b$ is too small, there is not enough information to predict future frames when $T_b + T_o$ is large, making it difficult to decide if a new frame should be selected.

\begin{table*}[t]
\caption{\textbf{Evaluation on Penn Action and Sub-JHDMB Dataset.} 
We achieve state-of-art performance on both datasets, using same pose model as \cite{nie2019dynamic}, but without any fine-tuning and using a small number of the key frames.} . 
\label{penn-jhmdb}
\centering
\begin{adjustbox}{width=1.0\linewidth}
\begin{tabular}{|c|c|c|c|c|c|c|c|c|c|c|c|}
\hline
\multicolumn{12}{|c|}{\textbf{Evaluation on Penn Action dataset}} \\ \hline
Method                             & FLOPs(G)      & Time(ms)    & Head                      & Sho.                     & Elb.                     & Wri.                     & Hip                      & Knee                              & Ank.                     & Avg.                      & Key frames(Std.) \\ \hline
Nie \etal\cite{xiaohan2015joint}                          & -             & -            & 64.2                      & 55.4                     & 33.8                     & 22.4                     & 56.4                     & 54.1                              & 48.0                     & 48.0                     &         N/A   \\ \hline
Iqal \etal\cite{iqbal2017pose}                         & -             & -            & 89.1                      & 86.4                     & 73.9                     & 73.0                     & 85.3                     & 79.9                              & 80.3                     & 81.1                     &     N/A       \\ \hline
Gkioxari \etal\cite{gkioxari2016chained}                     & -             & -            & 95.6                      & 93.8                     & 90.4                     & 90.7                     & 91.8                     & 90.8                              & 91.5                     & 91.9                     &       N/A     \\ \hline
Song \etal\cite{song2017thin}                        & -             & -            & 98.0                      & 97.3                     & 95.1                     & 94.7                     & 97.1                     & 97.1                              & 96.9                     & 96.8                     &       N/A     \\ \hline
Luo \etal\cite{luo2018lstm}                          & 70.98         & 25.0           & 98.9                      & 98.6                     & 96.6                     & 96.6                     & 98.2                     & 98.2                              & 97.5                     & 97.7                     &      N/A      \\ \hline
DKD(smallCPM) \cite{nie2019dynamic}                      & 9.96          & 12.0           & 98.4                      & 97.3                     & 96.1                     & 95.5                     & 97.0                     & 97.3                              & 96.6                     & 96.8                     &      N/A      \\ \hline
baseline \cite{simple-baseline}                      & 11.96          & 11.3           & 98.1                      & 98.2                     & 96.3                     & 96.4                     & 98.4                    & 97.5                              & 97.1                     & 97.4                     &     N/A       \\ \hline
DKD(Resnet50) \cite{nie2019dynamic}                      & 8.65          & 11.0           & 98.8                      & 98.7                     & 96.8                     & 97.0                     & 98.2                     & 98.1                              & 97.2                     & 97.8                     &     N/A       \\ \hline

\textbf{Ours(Resnet50)} & 5.37 & 6.8 & 98.7 & \textbf{98.7} & \textbf{97.0} & 95.3 & \textbf{98.8} & \textbf{98.7} & \textbf{98.6} & \textbf{98.0} & 17.5(4.9)   \\ \hline 

\textbf{Ours(Resnet34)} & \textbf{4.68} & \textbf{5.3} & 98.2 & 98.2 & 96.0 & 93.6 & 98.7 & 98.6 & 98.4 & 97.4 & \textbf{15.2(3.3)}   \\ \hline 

\end{tabular}
\end{adjustbox}
\bigskip
\centering
\begin{adjustbox}{width=1.0\linewidth}
\begin{tabular}{|c|c|c|c|c|c|c|c|c|c|c|c|}
\hline
\multicolumn{12}{|c|}{\textbf{Evaluation on sub-JHMDB dataset}}                                                                                                                                                                                                                                 \\ \hline
Methods         &FLOPs(G)      & Time(ms)                                           & Head                     & Sho.                     & Elbow                    & Wrist                    & Hip                      & Knee                     & Ankle & Avg.     &Key frames(Std.)                \\ \hline
Park \etal\cite{park2011n}       &- &-                                               & 79.0                     & 60.3                     & 28.7                     & 16.0                     & 74.8                     & 59.2                     & 49.3  & 52.5        &         N/A    \\ \hline
Nie \etal\cite{xiaohan2015joint}        &- &-                                                 & 83.3                     & 63.5                     & 33.8                     & 21.6                     & 76.3                     & 62.7                     & 53.1  & 55.7        &      N/A       \\ \hline
Iqal \etal\cite{iqbal2017pose}      &-    &-                                   & 90.3                     & 76.9                     & 59.3                     & 55.0                     & 85.9                     & 76.4                     & 73.0  & 73.8            &    N/A     \\ \hline
Song \etal\cite{song2017thin}         &- &-                                               & 97.1 & 95.7 & 87.5 & 81.6 & 98.0 & 92.7 & 89.8  & 92.1 &N/A \\ \hline
Luo \etal\cite{luo2018lstm}          &70.98 & 24.0                                             & 98.2                     & 96.5                     & 89.6                     & 86.0                     & 98.7                     & 95.6                     & 90.0  & 93.6                &  N/A   \\ \hline
DKD(Resnet50) \etal\cite{nie2019dynamic} &8.65    &- & 98.3 & 96.6 & 90.4 & 87.1 & 99.1 & 96.0 & 92.9  & 94.0  &N/A \\ \hline
baseline \etal\cite{simple-baseline} &11.96    &10.0 & 97.5 & 97.8 &91.1 & 86.0 & 99.6 & 96.8 & 92.6  & 94.4  &N/A \\ \hline

\textbf{Ours(Resnet50)}  &5.37    & 7.0 &  95.1 & 96.4 & \textbf{95.3} & \textbf{91.3} & 96.3 & 95.6 & 92.6    & \textbf{94.7}  & 17.8(1.4)\\ \hline  
\textbf{Ours(Resnet18)} &\textbf{4.68}    & \textbf{4.7} &  94.7 & 96.3 & 95.2 & 90.2 & 96.4 & 95.5 & \textbf{93.2}     & 94.5  & \textbf{16.3(1.8)}\\ \hline  
\end{tabular}
\end{adjustbox}
\end{table*}

\subsection{Comparison Against the State-of-Art}

 Comparisons against the state-of-art are reported in Table \ref{penn-jhmdb}. We report our performance using Resnet34 for Penn Action and Resnet18 for  Sub-JHMDB, and also using Resnet50, since  it is the backbone used by \cite{nie2019dynamic}. Our approach achieves the best performance and is 1.6X faster (6.8ms v.s 11ms) than the previous state-of-art 
 \cite{nie2019dynamic} for the Penn Action dataset, using an average of 17.5 key frames. Moreover, if we use our lightest model (Resnet34), our approach is 2X faster than \cite{nie2019dynamic}  with  a minor PCK degradation. For the sub-JHMDB dataset,  \cite{nie2019dynamic} did not provide running time and it is not open-sourced. Thus, we compare time against the  best available open sourced method \cite{luo2018lstm}. Our approach performed the best of all  methods, with a significant improvement on elbow (95.3\%) and wrist (91.3\%). For completeness,  we also compared against the baseline \cite{simple-baseline}, which is a frame-based method, on both datasets. We can observe that by applying our approach with the lightest model, we run more than 2X faster than \cite{simple-baseline} without any degradation in accuracy. 
 
 \subsection{{Robustness of Our Approach}}
 
 \begin{table}[t]
\caption{\textbf{Robustness Evaluation} }
\label{robust}
\centering
\begin{adjustbox}{width=1.0\linewidth}
\begin{tabular}{|r|c|c|c|c|c|c|c|}
\hline
\multicolumn{8}{|c|}{Perturbed frame ratio v.s average PCK score on sub-JHMBD}                                                                                                                                                                                                                 \\ \hline
Perturbed frames (\%)                            & 0                              & 10                             & 20                             & 30                             & 40                             & 50                             & 60                             \\ \hline
Illum.changes: \cite{simple-baseline}/Ours                    & 94.4/94.5                      & 94.0/94.2                      & 93.2/93.7                      & 92.3/93.0                      & 91.6/92.7                      & 90.9/92.3                      & 90.2/92.1                      \\ \hline
Blurring: \cite{simple-baseline}/Ours                         & 94.4/94.5                      & 92.6/93.4                      & 91.1/92.7                      & 89.9/91.7                      & 89.1/91.4                      & 88.5/91.2                      & 87.9/91.0                      \\ \hline
\multicolumn{1}{|r|}{Occlusions: \cite{simple-baseline}/Ours} & \multicolumn{1}{l|}{94.4/94.5} & \multicolumn{1}{l|}{92.8/94.0} & \multicolumn{1}{l|}{90.8/93.1} & \multicolumn{1}{l|}{89.3/92.1} & \multicolumn{1}{l|}{88.0/91.7} & \multicolumn{1}{l|}{86.5/91.3} & \multicolumn{1}{l|}{85.4/90.4} \\ \hline
\end{tabular}
\end{adjustbox}
\end{table}

We hypothesize that our approach can achieve better performance than previous approaches using fewer input frames because the network selects ``good" input frames, which are more robust when used with the frame-based method \cite{simple-baseline}. To better quantify this, we ran an experiment where we randomly partially occluded/blurred/changed illumination at random frames in the sub-JHMBD dataset. Table \ref{robust} shows that our approach (using ResNet18) is more robust to all of these perturbations when compared to \cite{simple-baseline}.

\section{Conclusion}
In this paper, we introduced a {\em key frame proposal} network (K-FPN) and a {\em human pose interpolation} module (HPIM) for efficient video based pose estimation. The proposed K-FPN can identify the dynamically informative frames from a video, which allows an image based pose estimation model to focus on only a few ``good" frames instead of the entire video. With a suitably learned pose dynamics-based dictionary, we show that the entire pose sequence can be recovered by the HPIM, using only the pose information from the  frames selected by the K-FPN.  The proposed method achieves  better (similar) accuracy than current state-of-art methods using 60\% ( 50\%) of the inference time.

\section*{Acknowledgements}
This work was supported  by NSF grants IIS--1814631 and ECCS--1808381;  and the Alert DHS Center of
Excellence under Award Number 2013-ST-061-ED0001. The views and conclusions contained in this document are those of the authors and should not be interpreted as necessarily representing the official policies, either expressed or implied, of the U.S. Department of Homeland Security.

\clearpage
\bibliographystyle{splncs04}
\bibliography{2841}

\begin{thebibliography}{10}
\providecommand{\url}[1]{\texttt{#1}}
\providecommand{\urlprefix}{URL }
\providecommand{\doi}[1]{https://doi.org/#1}

\bibitem{mpii}
Andriluka, M., Pishchulin, L., Gehler, P., Schiele, B.: 2d human pose
  estimation: New benchmark and state of the art analysis. In: IEEE Conference
  on Computer Vision and Pattern Recognition (CVPR) (June 2014)

\bibitem{belagiannis2017recurrent}
Belagiannis, V., Zisserman, A.: Recurrent human pose estimation. In: 2017 12th
  IEEE International Conference on Automatic Face \& Gesture Recognition (FG
  2017). pp. 468--475. IEEE (2017)

\bibitem{sparsely-labeled}
Bertasius, G., Feichtenhofer, C., Tran, D., Shi, J., Torresani, L.: Learning
  temporal pose estimation from sparsely-labeled videos. In: Wallach, H.,
  Larochelle, H., Beygelzimer, A., d\' Alch\'{e}-Buc, F., Fox, E., Garnett, R.
  (eds.) Advances in Neural Information Processing Systems 32, pp. 3027--3038.
  Curran Associates, Inc. (2019),
  \url{http://papers.nips.cc/paper/8567-learning-temporal-pose-estimation-from-sparsely-labeled-videos.pdf}

\bibitem{openpose}
Cao, Z., Hidalgo, G., Simon, T., Wei, S.E., Sheikh, Y.: Open{P}ose: realtime
  multi-person 2{D} pose estimation using {P}art {A}ffinity {F}ields. In: arXiv
  preprint arXiv:1812.08008 (2018)

\bibitem{charles2016personalizing}
Charles, J., Pfister, T., Magee, D., Hogg, D., Zisserman, A.: Personalizing
  human video pose estimation. In: Proceedings of the IEEE Conference on
  Computer Vision and Pattern Recognition. pp. 3063--3072 (2016)

\bibitem{chen2014articulated}
Chen, X., Yuille, A.L.: Articulated pose estimation by a graphical model with
  image dependent pairwise relations. In: Advances in Neural Information
  Processing Systems. pp. 1736--1744 (2014)

\bibitem{chu2016structured}
Chu, X., Ouyang, W., Li, H., Wang, X.: Structured feature learning for pose
  estimation. In: Proceedings of the IEEE Conference on Computer Vision and
  Pattern Recognition. pp. 4715--4723 (2016)

\bibitem{cristani2013human}
Cristani, M., Raghavendra, R., Del~Bue, A., Murino, V.: Human behavior analysis
  in video surveillance: A social signal processing perspective. Neurocomputing
   \textbf{100},  86--97 (2013)

\bibitem{gkioxari2016chained}
Gkioxari, G., Toshev, A., Jaitly, N.: Chained predictions using convolutional
  neural networks. In: European Conference on Computer Vision. pp. 728--743.
  Springer (2016)

\bibitem{Resnet}
He, K., Zhang, X., Ren, S., Sun, J.: Deep residual learning for image
  recognition. CoRR  \textbf{abs/1512.03385} (2015),
  \url{http://arxiv.org/abs/1512.03385}

\bibitem{flownet}
Ilg, E., Mayer, N., Saikia, T., Keuper, M., Dosovitskiy, A., Brox, T.: Flownet
  2.0: Evolution of optical flow estimation with deep networks. CoRR
  \textbf{abs/1612.01925} (2016), \url{http://arxiv.org/abs/1612.01925}

\bibitem{iqbal2017pose}
Iqbal, U., Garbade, M., Gall, J.: Pose for action-action for pose. In: 2017
  12th IEEE International Conference on Automatic Face \& Gesture Recognition
  (FG 2017). pp. 438--445. IEEE (2017)

\bibitem{poseTrack}
Iqbal, U., Milan, A., Gall, J.: Pose-track: Joint multi-person pose estimation
  and tracking. CoRR  \textbf{abs/1611.07727} (2016),
  \url{http://arxiv.org/abs/1611.07727}

\bibitem{Jhuang:ICCV:2013}
Jhuang, H., Gall, J., Zuffi, S., Schmid, C., Black, M.J.: Towards understanding
  action recognition. In: International Conf. on Computer Vision (ICCV). pp.
  3192--3199 (Dec 2013)

\bibitem{PiPaf}
Kreiss, S., Bertoni, L., Alahi, A.: Pifpaf: Composite fields for human pose
  estimation. In: The IEEE Conference on Computer Vision and Pattern
  Recognition (CVPR) (June 2019)

\bibitem{lin2010augmented}
Lin, H.Y., Chen, T.W.: Augmented reality with human body interaction based on
  monocular 3d pose estimation. In: International Conference on Advanced
  Concepts for Intelligent Vision Systems. pp. 321--331. Springer (2010)

\bibitem{3Dlstm}
Lin, M., Lin, L., Liang, X., Wang, K., Cheng, H.: Recurrent 3d pose sequence
  machines. CoRR  \textbf{abs/1707.09695} (2017),
  \url{http://arxiv.org/abs/1707.09695}

\bibitem{featurepriamid}
Lin, T., Doll{\'{a}}r, P., Girshick, R.B., He, K., Hariharan, B., Belongie,
  S.J.: Feature pyramid networks for object detection. CoRR
  \textbf{abs/1612.03144} (2016), \url{http://arxiv.org/abs/1612.03144}

\bibitem{DYAN}
Liu, W., Sharma, A., Camps, O.I., Sznaier, M.: {DYAN:} {A} dynamical atoms
  network for video prediction. CoRR  \textbf{abs/1803.07201} (2018),
  \url{http://arxiv.org/abs/1803.07201}

\bibitem{luo2018lstm}
Luo, Y., Ren, J., Wang, Z., Sun, W., Pan, J., Liu, J., Pang, J., Lin, L.:
  {LSTM} pose machines. In: Proceedings of the IEEE Conference on Computer
  Vision and Pattern Recognition. pp. 5207--5215 (2018)

\bibitem{Hourglass}
Newell, A., Yang, K., Deng, J.: Stacked hourglass networks for human pose
  estimation. CoRR  \textbf{abs/1603.06937} (2016),
  \url{http://arxiv.org/abs/1603.06937}

\bibitem{nie2018mula}
Nie, X., Feng, J., Yan, S.: Mutual learning to adapt for joint human parsing
  and pose estimation. In: ECCV (2018)

\bibitem{nie2019dynamic}
Nie, X., Li, Y., Luo, L., Zhang, N., Feng, J.: Dynamic kernel distillation for
  efficient pose estimation in videos. In: The IEEE International Conference on
  Computer Vision (ICCV) (October 2019)

\bibitem{PapandreouZKTTB17}
Papandreou, G., Zhu, T., Kanazawa, N., Toshev, A., Tompson, J., Bregler, C.,
  Murphy, K.P.: Towards accurate multi-person pose estimation in the wild. CoRR
   \textbf{abs/1701.01779} (2017), \url{http://arxiv.org/abs/1701.01779}

\bibitem{park2011n}
Park, D., Ramanan, D.: N-best maximal decoders for part models. In: 2011
  International Conference on Computer Vision. pp. 2627--2634. IEEE (2011)

\bibitem{park2008understanding}
Park, S., Trivedi, M.M.: Understanding human interactions with track and body
  synergies (tbs) captured from multiple views. Computer Vision and Image
  Understanding  \textbf{111}(1),  2--20 (2008)

\bibitem{pfister2015flowing}
Pfister, T., Charles, J., Zisserman, A.: Flowing convnets for human pose
  estimation in videos. In: Proceedings of the IEEE International Conference on
  Computer Vision. pp. 1913--1921 (2015)

\bibitem{pishchulin2013strong}
Pishchulin, L., Andriluka, M., Gehler, P., Schiele, B.: Strong appearance and
  expressive spatial models for human pose estimation. In: Proceedings of the
  IEEE International Conference on Computer Vision. pp. 3487--3494 (2013)

\bibitem{shotton2011real}
Shotton, J., Fitzgibbon, A., Cook, M., Sharp, T., Finocchio, M., Moore, R.,
  Kipman, A., Blake, A.: Real-time human pose recognition in parts from single
  depth images. In: CVPR 2011. pp. 1297--1304. Ieee (2011)

\bibitem{simonyan2014two}
Simonyan, K., Zisserman, A.: Two-stream convolutional networks for action
  recognition in videos. In: Advances in neural information processing systems.
  pp. 568--576 (2014)

\bibitem{song2017thin}
Song, J., Wang, L., Van~Gool, L., Hilliges, O.: Thin-slicing network: A deep
  structured model for pose estimation in videos. In: Proceedings of the IEEE
  Conference on Computer Vision and Pattern Recognition. pp. 4220--4229 (2017)

\bibitem{Tang_dlcm}
Tang, W., Yu, P., Wu, Y.: Deeply learned compositional models for human pose
  estimation. In: The European Conference on Computer Vision (ECCV) (September
  2018)

\bibitem{TempoBai}
{Tempo}, R., {Bai}, E.W., {Dabbene}, F.: Probabilistic robustness analysis:
  explicit bounds for the minimum number of samples. In: Proceedings of 35th
  IEEE Conference on Decision and Control. vol.~3, pp. 3424--3428 vol.3 (Dec
  1996)

\bibitem{deeppose}
Toshev, A., Szegedy, C.: Deeppose: Human pose estimation via deep neural
  networks. In: 2014 IEEE Conference on Computer Vision and Pattern
  Recognition. pp. 1653--1660 (June 2014). \doi{10.1109/CVPR.2014.214}

\bibitem{wei2016convolutional}
Wei, S.E., Ramakrishna, V., Kanade, T., Sheikh, Y.: Convolutional pose
  machines. In: Proceedings of the IEEE Conference on Computer Vision and
  Pattern Recognition. pp. 4724--4732 (2016)

\bibitem{simple-baseline}
Xiao, B., Wu, H., Wei, Y.: Simple baselines for human pose estimation and
  tracking. CoRR  \textbf{abs/1804.06208} (2018),
  \url{http://arxiv.org/abs/1804.06208}

\bibitem{xiaohan2015joint}
Xiaohan~Nie, B., Xiong, C., Zhu, S.C.: Joint action recognition and pose
  estimation from video. In: Proceedings of the IEEE Conference on Computer
  Vision and Pattern Recognition. pp. 1293--1301 (2015)

\bibitem{quantization}
Yang, J., Shen, X., Xing, J., Tian, X., Li, H., Deng, B., Huang, J., Hua, X.s.:
  Quantization networks. In: The IEEE Conference on Computer Vision and Pattern
  Recognition (CVPR) (June 2019)

\bibitem{yang2017pyramid}
Yang, W., Li, S., Ouyang, W., Li, H., Wang, X.: Learning feature pyramids for
  human pose estimation. In: arXiv preprint arXiv:1708.01101 (2017)

\bibitem{yang2016end}
Yang, W., Ouyang, W., Li, H., Wang, X.: End-to-end learning of deformable
  mixture of parts and deep convolutional neural networks for human pose
  estimation. In: Proceedings of the IEEE Conference on Computer Vision and
  Pattern Recognition. pp. 3073--3082 (2016)

\bibitem{Yang&Ramanan}
Yang, Y., Ramanan, D.: Articulated pose estimation with flexible
  mixtures-of-parts. In: CVPR 2011. pp. 1385--1392 (June 2011).
  \doi{10.1109/CVPR.2011.5995741}

\bibitem{fastpose}
Zhang, F., Zhu, X., Ye, M.: Fast human pose estimation. In: The IEEE Conference
  on Computer Vision and Pattern Recognition (CVPR) (June 2019)

\bibitem{penn_action}
{Zhang}, W., {Zhu}, M., {Derpanis}, K.G.: From actemes to action: A
  strongly-supervised representation for detailed action understanding. In:
  2013 IEEE International Conference on Computer Vision. pp. 2248--2255 (Dec
  2013). \doi{10.1109/ICCV.2013.280}

\end{thebibliography}
\end{document}